\documentclass[10pt,twocolumn,letterpaper]{article}

\usepackage[pagenumbers]{cvpr} 

\usepackage{pifont}
\newcommand{\cmark}{\ding{51}} 
\newcommand{\xmark}{\ding{55}} 
\usepackage{multirow}
\usepackage{float}
\usepackage[table]{xcolor}

\definecolor{cvprblue}{rgb}{0.2,0.2,0.8} 
\usepackage[pagebackref,breaklinks,colorlinks,allcolors=cvprblue]{hyperref}

\def\paperID{*****} 
\def\confName{CVPR}
\def\confYear{2026}

\title{DRG-Font: Dynamic Reference-Guided Few-shot Font Generation via Contrastive Style-Content Disentanglement}

\author{
Rejoy Chakraborty$^{1*}$ \qquad
Prasun Roy$^{1*}$ \qquad
Saumik Bhattacharya$^{2}$ \qquad
Umapada Pal$^{1}$ \vspace{0.5em}\\
$^{1}$Indian Statistical Institute Kolkata \qquad $^{2}$Indian Institute of Technology Kharagpur \vspace{0.5em}\\
{\tt\small\textbf{\url{https://rejoycs.github.io/drg-font}}}
}

\begin{document}


\twocolumn[{%
\renewcommand\twocolumn[1][]{#1}%
\maketitle
\vspace{-3.6em}
\begin{center}
    \includegraphics[width=\textwidth]{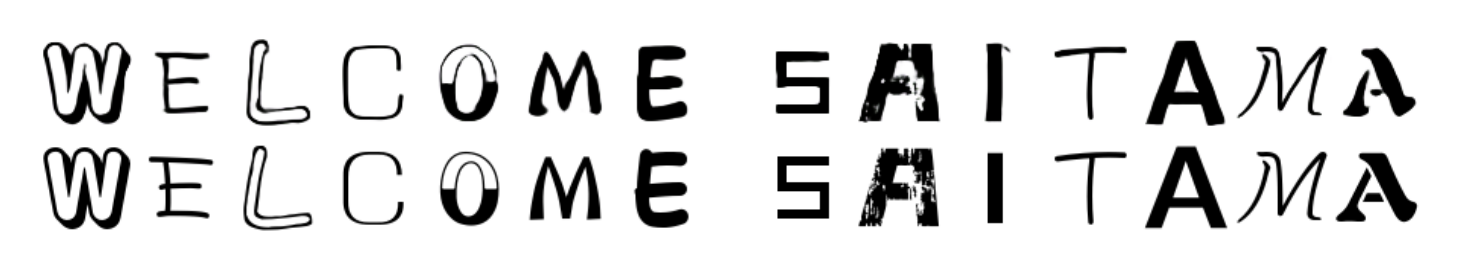}
    \vspace{-2.0em}
    \captionof{figure}{Examples of generated instances using the proposed DRG-Font. The top row shows the generated glyphs, and the bottom row shows the corresponding ground truth glyphs.}
    \label{fig:teaser}
\end{center}
}]
\let\thefootnote\relax\footnotetext{$^*$Equal contribution.}

\begin{abstract}
Few-shot Font Generation aims to generate stylistically consistent glyphs from a few reference glyphs. However, capturing complex font styles from a few exemplars remains challenging, and the existing methods often struggle to retain discernible local characteristics in generated samples. This paper introduces DRG-Font, a contrastive font generation strategy that learns complex glyph attributes by decomposing style and content embedding spaces. For optimal style supervision, the proposed architecture incorporates a Reference Selection (RS) Module to dynamically select the best style reference from an available pool of candidates. The network learns to decompose glyph attributes into style and shape priors through a Multi-scale Style Head Block (MSHB) and a Multi-scale Content Head Block (MCHB). For style adaptation, a Multi-Fusion Upsampling Block (MFUB) produces the target glyph by combining the reference style prior and target content prior. The proposed method demonstrates significant improvements over state-of-the-art approaches across multiple visual and analytical benchmarks.
\end{abstract}

\section{Introduction}
\label{sec:intro}

Few-shot Font Generation (FFG) is a conditional Image-to-Image (I2I) translation technique. In FFG, the objective is to generate a character glyph image that adopts a target font style using a few style reference images. In conventional scenarios, glyph images are rendered using TrueType Font (TTF) files, which encode the geometric and stylistic attributes of a font. However, generating the corresponding TTF file by estimating the geometric and style properties using a few style reference images remains a challenging problem. FFG addresses this limitation by transferring the style from the provided reference images to a given character structure without explicitly reconstructing the underlying TTF file representation.

With recent advances of deep generative models, such as Generative Adversarial Networks (GAN)~\cite{goodfellow2014generative} and diffusion models~\cite{ho2020denoising}, numerous studies have explored style transfer tasks. Furthermore, several works have focused on imposing the desired style properties on character glyphs. Early approaches attempted to transfer style from one character to another using I2I translation methods~\cite{isola2017image, liu2019few, roy2020stefann}. However, these approaches were limited to a set of mappings between predefined domains. Subsequently, style-content disentanglement-based methods~\cite{xie2021dg, park2021multiple} were introduced to decompose the underlying style and content representations from respective input images, which were then combined to generate the target glyph. More recent methods~\cite{xie2021dg, zeng2025few} incorporate structure-aware approaches that decompose characters into predefined strokes to improve generation quality.

Despite the advantages of structural decomposition, such methods remain limited to script in which characters can be represented using a fixed set of strokes. Moreover, such decompositions are script-dependent, and for scripts without well-defined stroke decomposition rules, multiple valid user-defined strategies may exist. Consequently, these approaches cannot provide a generalized way of capturing complex style features across different scripts. It is also worth noting that diffusion-based methods generally produce sharp, high-quality results. However, substantial computational overheads limit their practicality in many deployment scenarios.

To address the aforementioned limitations, the proposed pipeline incorporates a Reference Selection (RS) Module that selects the optimal style reference from a pool of available candidates by measuring a similarity metric. The generator network consists of an encoder and a decoder, which takes the selected style reference along with the content reference as inputs to produce the target character in the desired font style. The encoder decomposes both inputs into their respective style and content embedding spaces. Subsequently, the decoder performs a cross-embedding fusion to generate the target glyph. The style and content embedding spaces are learned using a contrastive learning strategy. Additionally, the hybrid learning objective utilizes a latent reconstruction loss to ensure high-quality latent representations and a discriminator-based loss for adversarial guidance.

\noindent
\textbf{Contributions:} The key contributions of the proposed DRG-Font can be summarized as follows.

\begin{itemize}
    \item The proposed method introduces a novel dynamic style reference selection strategy through the RS Module that significantly improves the ability to retain glyph characteristics in generated samples by selecting the optimal style reference from a set of candidates.

    \item The method proposes a novel contrastive strategy for decomposing the style and content embedding spaces, followed by a cross-embedding fusion to generate the target glyph. The network is optimized using a hybrid objective consisting of contrastive, reconstruction, and adversarial loss components.

    \item The proposed method generalizes well to different scripts (Latin and Chinese). It outperforms the existing state-of-the-art (SOTA) font style generation techniques in both qualitative and quantitative evaluations.
\end{itemize}

\noindent
The rest of the paper is organized as follows. Section~\ref{sec:related} provides a brief overview of the existing literature; the proposed methodology is discussed in detail in Section~\ref{sec:method}, followed by results and analysis in Section~\ref{sec:results}; finally, the concluding remarks are discussed in Section~\ref{sec:conclusion}.

\begin{figure*}[t]
    \centering
    \includegraphics[width=0.9\linewidth]{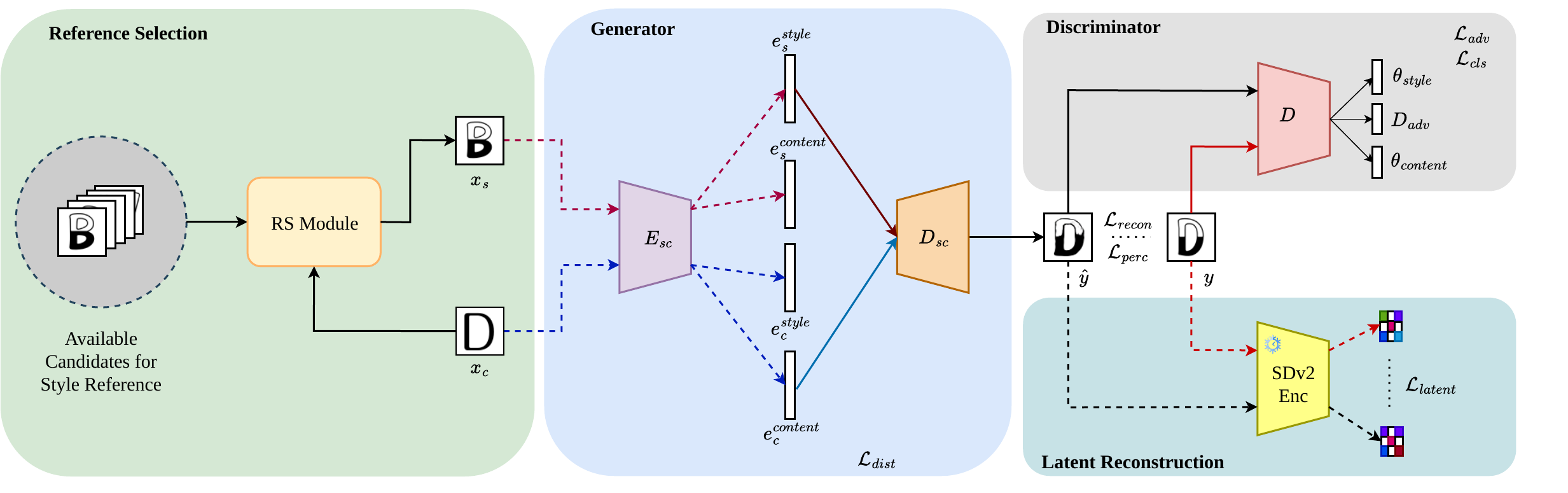}
    \caption{An overview of the proposed DRG-Font. The initial reference selection is performed by the RS Module, which finds the optimal style reference $x_s$ from an available set of candidates, such that $x_s$ is structurally closest to the given content reference $x_c$. The Generator network independently encodes $x_s$ and $x_c$ by using $E_{sc}$, producing the style and content embeddings from both references. The target glyph $\hat{y}$ is generated by $D_{sc}$ through a cross-fusion of the embeddings. The discriminator network enforces adversarial guidance by auxiliary style and content classification. The Stable Diffusion Encoder (SDv2 Enc)~\cite{rombach2022high} is used to ensure high-quality latent reconstruction.}
    \label{fig:workflow}
\end{figure*}

\section{Related Work}
\label{sec:related}

\subsection{Image-to-Image (I2I) Translation}
In I2I translation, a transformation function learns to translate content from the source domain to the target domain. Following the introduction of GAN~\cite{goodfellow2014generative}, several GAN-based I2I methods~\cite{yi2017dualgan, choi2018stargan} have been introduced. Pix2Pix~\cite{isola2017image} was one of the earliest approaches to formulate the concept of conditional GAN with paired data. Similarly, DualGAN~\cite{yi2017dualgan} uses unlabeled data pairs for unsupervised I2I translation. In CycleGAN~\cite{zhu2017unpaired}, a circular consistency loss is used for unsupervised I2I translation. Using the style-content pair as a source, FUNIT~\cite{liu2019few} performs the I2I translation task by mixing the embeddings of style and content features using AdaIN~\cite{huang2017arbitrary}. Later, diffusion-based methods~\cite{parmar2023zero, li2023bbdm, tumanyan2023plug} have been proposed to handle a more diverse and complex range of data.

\subsection{Few-shot Font Generation (FFG)}
FFG is essentially a conditional I2I translation that aims to generate a character glyph with a specified font style using a few style references as observed exemplars. Most early works~\cite{zhou2011easy, campbell2014learning, phan2015flexyfont} are generally based on structural properties such as strokes and radicals. However, these classical approaches show drastic limitations for complex artistic styles. In recent years, deep generative networks~\cite{yang2017awesome} have demonstrated significant improvements for font generation. STEFFAN~\cite{roy2020stefann} proposed the first scene text editing technique by introducing FANNET, a character-level adaptive font style generation network. MC-GAN~\cite{azadi2018multi} performs artistic glyph generation from a few observations, producing an entire set of characters in a single pass. Later studies have introduced style-content disentanglement strategies to decompose the reference attributes into separate style and content embedding spaces, which are subsequently fused to generate the target. DG-Font~\cite{xie2021dg} used a Feature Deformation Skip Connection module to apply deformable convolutions~\cite{zhu2019deformable} to capture low-level geometric variations between fonts. MX-Font~\cite{park2021multiple} introduced a multi-headed architecture composed of multiple localized experts and a generator, where each expert aims to capture distinct sub-structures of a glyph, enforced using HSIC~\cite{greenfeld2020robust}. FS-Font~\cite{tang2022few} comprises a SAM module that constructs the Query (Q), Key (K), and Value (V) triplet from the extracted features of the style and content encoders to learn the correspondence between the style reference and content features. MA-Font~\cite{qiu2024ma} incorporated a multi-level adaptation mechanism of style features into the content feature. DA-Font~\cite{chen2025font} introduces a dual-attention framework that leverages component-aware and relation-aware attention to enhance structural consistency and visual style fidelity.

In recent studies, CLIP~\cite{radford2021learning} embedding has been widely used on the segmented glyph strokes. CLIP-Font~\cite{xiong2024clip} highlights informative regions via contrastive learning and enforces content consistency by maximizing the cosine similarity between the text-image embeddings obtained through CN-CLIP~\cite{yang2022chinese}. SPH-Font~\cite{zeng2025few} incorporates a hierarchical representation learning scheme, using a Stroke Prompt (SP) module constructed by fine-tuning an IT-CLIP model. In contrast to stroke components decomposition-based approaches~\cite{sun2017learning, tang2022few, yao2024vq, chen2025font}, Patch-Font~\cite{memon2025patch} learns patch-level style representations from reference glyphs to synthesize new characters while preserving fine-grained stroke structures. FontDiffuser~\cite{yang2024fontdiffuser} proposes a one-shot generation technique using a conditional denoising diffusion model. It introduces multi-scale content aggregation to preserve fine stroke details and a style contrastive learning objective to enforce style consistency with only a single reference. Diff-Font~\cite{he2024diff} also uses a one-shot conditional diffusion architecture utilizing both decomposed components and strokes as conditions.

\section{Proposed Methodology}
\label{sec:method}

The proposed pipeline aims to disentangle the style and content feature spaces from respective reference images (style reference and content reference), followed by performing cross-style-content feature fusion to generate the target glyph.

Considering a set of $N$ fonts (styles) $\mathcal{F} = \{f_1, f_2, ..., f_N\}$, where each font style contains a set of $M$ characters/contents (will be used interchangeably) $\mathcal{C} = \{c_1, c_2, ... , c_M\}$. For notational ease, $x_{a,b}$ denotes a glyph image with font style $f_a \in \mathcal{F}$ of character $c_b \in \mathcal{C}$. Given a style reference image $x_s = x_{a,\star} \in f_a \times \mathcal{C}$ for target style $f_a$ of any character and a content reference image $x_c = x_{\star,b} \in \mathcal{F} \times c_b$ for the target character $c_b$ of any font, the font generation pipeline aims to generate $\hat{y} = x_{a,b}$. However, in practice~\cite{xie2021dg, yang2024fontdiffuser, chen2025font}, a fixed standard font style $f_t$, that can represent generic character structures across a wide variety of fonts, is used for the content reference, i.e., $x_c = x_{t,b}$ for the target character $c_b$.

The proposed method introduces a \textbf{R}eference \textbf{S}election (\textbf{RS}) \textbf{Module}, which selects the optimal style reference $x_s$ based on a dynamic selection criterion to improve the generation quality. After selection, $x_s$ and $x_c$ independently pass through the generator network \textbf{$G$}, which consists of the style-content encoder $E_{sc}$ and decoder $D_{sc}$. For each reference, $E_{sc}$ produces a pair of style and content embeddings, denoted as $\{e^{style}_s, e^{content}_s\}$ and $\{e^{style}_c, e^{content}_c\}$ for $x_s$ and $x_c$, respectively. During decoding, $D_{sc}$ performs a cross-feature fusion between $e^{style}_s$ and $e^{content}_c$ to generate $\hat{y}$. For high-quality generation, a discriminator network $D$ provides adversarial supervision, and a Stable Diffusion ~\cite{rombach2022high} encoder SDv2 Enc ensures rich latent reconstruction. The following subsections discuss individual components and the training objective of the proposed network. Figure \ref{fig:workflow} illustrates an overview of the DRG-Font architecture.

\begin{figure*}[t]
    \centering
    \includegraphics[width=\linewidth]{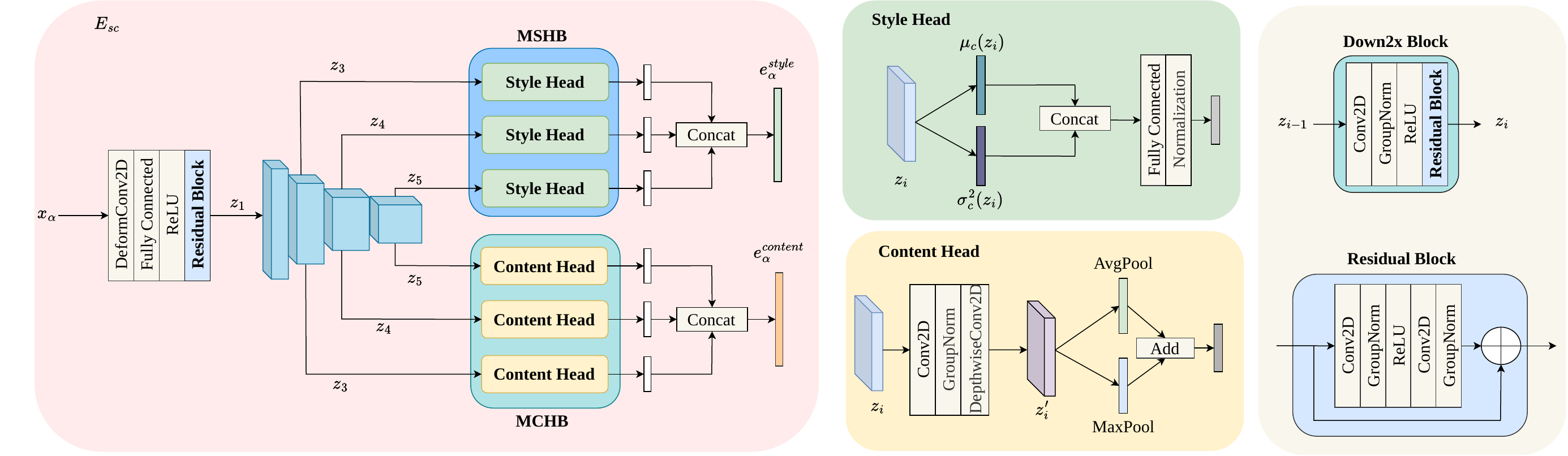}
    \caption{Architecture of the proposed Style-Content Encoder $E_{sc}$. The input image $x_\alpha$ passes through four consecutive Down2x Blocks. The resulting feature maps then pass through the Multiscale Style Head Block (MSHB) and the Multiscale Content Head Block (MCHB) in parallel. Both MSHB and MCHB encode three latent embeddings using three different heads, which are finally concatenated to produce the style embedding $e^ {style}_\alpha$ (from MSHB) and the content embedding $e^ {content}_\alpha$ (from MCHB).}
    \label{fig:encoder}
\end{figure*}

\subsection{Reference Selection}
While generating the target character with the desired style, a style reference that is structurally closer to the target character acts as a better style reference than a randomly chosen one. Therefore, for each character, there is a specific preference ordering of other characters. The proposed RS Module builds this preference table by using a stroke matching similarity measure.

To generate a target character $\hat{y} = x_{a,b}$, given the content reference image $x_c = x_{t,b}$, and a set of candidate observations $\mathcal{O}_a$ ($|\mathcal{O}_a| = M' \text{ and } M' < M$), the structural similarity score between each pair $\{x_c, x_{a,b'}\}$; $x_{a,b'} \in \mathcal{O}_a$, and $b'\neq b$, is computed using a \textbf{S}troke \textbf{M}atching \textbf{C}omparator (\textbf{SMC}) \textbf{Module}. The candidate $x_{a,b'}$ having the highest similarity score is selected as $x_s$.

\subsubsection{\textbf{Stroke Matching Comparator (SMC) Module:}}
To capture better structural similarity, the intrinsic topological and geometric properties of two given characters $x_c$ and $x_{a,b'}$ are measured, providing a finer similarity analysis. At first, a skeletonization~\cite{mahmoud1991skeletonization} operation is performed on $x_c$ and $x_{a,b'}$. For simplicity, the skeletonized images of $x_c$ and $x_{a,b'}$ are labeled as $A$ and $B$, respectively. Considering $A$ and $B$ as graphs, pixels having a degree of 1 or a degree more than 2 are considered as \textbf{`Salient Points'}.

The skeleton is decomposed into strokes by traversing paths between detected salient points. A stroke is defined as a maximal connected skeleton path that starts and ends at nodes without passing through intermediate salient points. Each stroke is represented as an ordered sequence of points along the skeleton. From this sequence, a descriptor is extracted using three components: \textbf{(1)} \textit{normalized stroke length}~\cite{bhattacharya2018sub}, computed as the sum of Euclidean distances between consecutive points; \textbf{(2)} \textit{average curvature}~\cite{lowe2004distinctive}, estimated from the change in orientation between successive stroke segments, where the orientation of each segment is computed over the coordinate differences of consecutive points; and \textbf{(3)} \textit{orientation distribution}~\cite{dalal2005histograms}, represented by a normalized histogram of segment orientations using 8 bins over the range $[-\pi, \pi]$. These features capture both global geometric properties and local directional variations of the stroke.

For the given two images, $A$ and $B$ with the descriptor sets of individual strokes $\mathcal{D}^A$ and $\mathcal{D}^B$, the pairwise cosine similarity is computed as $\mathbf{S}_{wv} = \operatorname{CosineSim}(\mathbf{d}_w^A, \mathbf{d}_v^B)$, where $\mathbf{d}_w^A \in \mathcal{D}^A$, $\mathbf{d}_v^B \in \mathcal{D}^B$, $1\leq w \leq |\mathcal{D}^A|$, $1\leq v \leq |\mathcal{D}^B|$, and $\operatorname{CosineSim}(\cdot,\cdot)$ indicates the cosine similarity between two vectors. To handle an uneven number of strokes for the characters, the final similarity score is defined as

{\small
\[
\text{Sim}(A,B)
=
\frac{1}{2}
\left(
\frac{1}{|\mathcal{D}^A|}
\sum_{w}
\max_v \mathbf{S}_{wv}
+
\frac{1}{|\mathcal{D}^B|}
\sum_{v}
\max_w \mathbf{S}_{wv}
\right).
\]
}

\subsection{\texorpdfstring{Generator Network $(G)$}{Generator Network}}
The proposed generator network consists of two components, $E_{sc}$ and $D_{sc}$. The Style-Content Encoder $E_{sc}$ extracts the style and content embeddings $e^{style}_\alpha, e^{content}_\alpha \in \mathbb{R}^{d}$ from an image $x_{\alpha}$, $\alpha \in \{s,c\}$. The Style-Content Decoder $D_{sc}$ generates the target glyph image $\hat{y}$ from the latent pair $\{e^{style}_s, e^{content}_c\}$.

\subsubsection{\texorpdfstring{\textbf{Style-Content Encoder $(E_{sc})$:}}{Style-Content Encoder}}
Given a reference image $x_{\alpha} \in \mathbb{R}^{C \times H \times W}$, a sequence of deformable convolution~\cite{zhu2019deformable}, group normalization, and ReLU activation is applied on $x_{\alpha}$ to produce  $z_1 \in \mathbb{R}^{C' \times H \times W}$. The deformable convolution helps to attain a better geometric invariance than traditional convolution. The resulting feature map $z_1$ passes through four consecutive \textbf{Down2x Blocks}, downsampling the feature space from $C_{i-1} \times H_{i-1} \times W_{i-1}$ to $C_i \times H_i \times W_i$, where $H_i = \frac{H_{i-1}}{2}, ~W_i = \frac{W_{i-1}}{2}, ~C_i = 2C_{i-1}$, and $H_0 = H, ~W_0 = W, ~C_0 = C$. Each Down2x Block incorporates a sequence of convolution, group normalization, and ReLU activation, followed by a \textbf{Residual Block}. The downsampling produces four latent feature maps $\{z_i \in \mathbb{R}^{C_i \times H_i \times W_i} ~|~ 2<=i<=5\}$, with progressively downscaled spatial resolutions. The last three latents $\{z_3, z_4, z_5\}$ are passed through the \textbf{M}ultiscale \textbf{S}tyle \textbf{H}ead \textbf{B}lock \textbf{(MSHB)} and the \textbf{M}ultiscale \textbf{C}ontent \textbf{H}ead \textbf{B}lock \textbf{(MCHB)} in parallel to produce the style and content embeddings, respectively. Figure~\ref{fig:encoder} shows the architecture of the Style-Content Encoder $E_{sc}$.

The MSHB consists of three style heads, where each style head performs a style projection on one of the $\{z_i ~|~ 3 \leq i \leq 5\}$ independently. A style head first computes the channel-wise mean $\mu_c(z_i) \in \mathbb{R}^{C_i}$ and variance $\sigma^2_c(z_i) \in \mathbb{R}^{C_i}$ from $z_i$. The concatenated vector $\langle\mu_c(z_i), \sigma^2_c(z_i)\rangle \in \mathbb{R}^{2C_i}$ represents a statistical measure of the style features and is projected to a style embedding $e^{style}_{\alpha, i} \in \mathbb{R}^{\frac{d}{3}}$. These individual style embeddings are used in the decoder $D_{sc}$ to adapt multiscale style representations. The final style embedding from MSHB is constructed by concatenating all the style representations from individual heads $e^{style}_\alpha = \langle e^{style}_{\alpha,3}, e^{style}_{\alpha,4}, e^{style}_{\alpha,5} \rangle \in \mathbb{R}^d$.

Similarly, MCHB consists of three content heads for $z_3$, $z_4$, and $z_5$. A content head first performs a feature space projection using convolution, group normalization, and depthwise separable convolution~\cite{chollet2017xceptiondeeplearningdepthwise} on $z_i$ to produce $z'_i \in \mathbb{R}^{\frac{d}{3} \times H_i \times W_i}$. Subsequently, a content embedding $e^{content}_{\alpha, i} \in \mathbb{R}^{\frac{d}{3}}$ is computed by aggregating the embeddings obtained by independently applying average pooling and max pooling on $z'_i$. The final content embedding from MCHB is a concatenation of all the feature vectors from individual heads $e^{content}_\alpha = \langle e^{content}_{\alpha,3}, e^{content}_{\alpha,4}, e^{content}_{\alpha,5} \rangle \in \mathbb{R}^d$. The decoder $D_{sc}$ uses the encoded features $e^{content}_\alpha$ to adapt the structural representation.

\subsubsection{\texorpdfstring{\textbf{Style-Content Decoder $(D_{sc})$:}}{Style-Content Decoder}}
\begin{figure}[t]
    \centering
    \includegraphics[width=\linewidth]{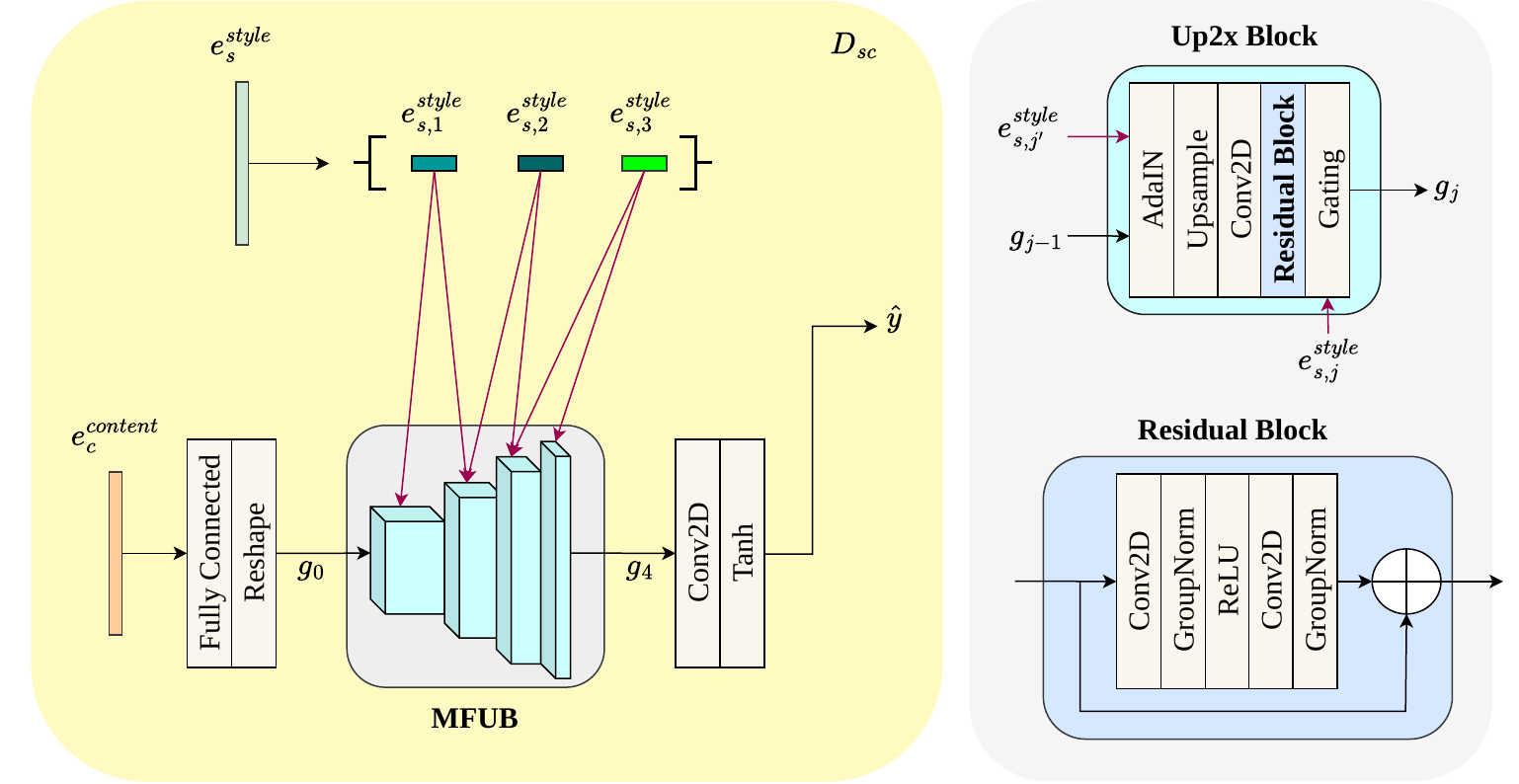}
    \caption{Architecture of the proposed Style-Content Decoder $D_{sc}$. The network uses a Multi-Fusion Upsampling Block (MFUB) to generate the target image $\hat{y}$ by combining the multiscale style embeddings $e^{style}_s= \langle e^{style}_{s,1}, e^{style}_{s,2}, e^{style}_{s,3} \rangle$ from MSHB and content embedding $e^{content}_c$ from MCHB.}
    \label{fig:decoder}
\end{figure}

Given the style-content pair $\{e^{style}_s$, $e^{content}_c\}$, the decoder $(D_{sc})$ generates the target glyph $\hat{y}$ using the font style features encoded in $e^{style}_s$ and the structural attributes of the target character encoded in $e^{content}_c$. Initially, a low-resolution latent $g_0 \in \mathbb{R}^{C_0 \times H_0 \times W_0}$ containing the structural features is produced from $e^{content}_c$, where $H_0 = \frac{H}{16}$, $W_0 = \frac{W}{16}$, $C_0 = 16C$. The vector $e^{style}_s$ consists of three embeddings $\{e^{style}_{s,1}, e^{style}_{s,2}, e^{style}_{s,3}\}$ of equal length, obtained from three independent style heads of MSHB to encode the target style information at multiple scales. $D_{sc}$ uses a \textbf{M}ulti-\textbf{F}usion \textbf{U}psampling \textbf{B}lock \textbf{(MFUB)} that progressively projects the multiscale style embeddings on the target character structure to produce $\hat{y}$. Figure~\ref{fig:decoder} shows the architecture of the proposed Style-Content Decoder $D_{sc}$.

The MFUB performs upsampling using four consecutive \textbf{Up2x Blocks}. In $j^{th}$ Up2x Block, $g_{j-1}$ first adapts to the style feature $e^{style}_{s, j'}$ using $\operatorname{AdaIN}$~\cite{huang2017arbitrary} to produce $g_j^{ad} \in \mathbb{R}^{C_{j-1} \times H_{j-1} \times W_{j-1}}$; where $j'=1$ when $j=1$, else $j'=j-1$. The feature space $g_j^{ad}$ is upsampled using bilinear interpolation and then forwarded through a convolution layer and a \textbf{Residual Block}, producing $g_j^{up} \in \mathbb{R}^{C_j \times H_j \times W_j}$, where $H_j = 2H_{j-1}$, $W_j = 2W_{j-1}$, $C_j = \frac{C_{j-1}}{2}$. Furthermore, a style-conditioned gating mechanism~\cite{perez2018film} is applied on the first three Up2x Blocks, yielding $g_j \in \mathbb{R}^{C_j \times H_j \times W_j}$. A gating vector is obtained from the style embedding $e^{style}_{s, j}$ on $g_j^{up}$ through a linear projection, and sigmoid activation modulated via channel-wise multiplication. The target image $\hat{y}\in \mathbb{R}^{C \times H \times W}$ is generated from the final latent $g_4$.

\subsection{\texorpdfstring{Discriminator Network $(D)$}{Discriminator Network}}
DRG-Font adopts the PatchGAN~\cite{isola2017image} architecture as a multi-task discriminator network $(D)$ to provide adversarial guidance by differentiating between real and generated glyphs. Additionally, $D$ provides auxiliary supervision~\cite{odena2017conditional} for style and content classification. Given an input image $x' \in \mathbb{R}^{C \times H \times W}$, a shared convolutional backbone with spectral normalization and LeakyReLU activation first extracts the hierarchical features of dimension ${C_D \times H_D \times W_D}$. An adversarial head estimates a patchwise binary class label map $D_{adv}(x') \in \mathbb{R}^{1 \times H_D \times W_D}$ from the hierarchical features to evaluate realism at multiple spatial resolutions. Furthermore, a classification head performs global average pooling on the same hierarchical feature space, followed by two independent branches with spectral normalization and fully connected layers to predict style and content classification logits $\{\theta_{style}(x') \in \mathbb{R}^{N}, \theta_{content}(x') \in \mathbb{R}^{M}\}$.

\subsection{Training Objective}
DRG-Font uses a hybrid objective consisting of six loss components for spatial reconstruction, perceptual quality, adversarial supervision, output classification, contrastive disentanglement, and latent reconstruction.

\subsubsection{\textbf{Reconstruction Loss:}}
The pixel-wise reconstruction loss between the generated image $\hat{y}$ and the ground truth $y$ is estimated as the $\ell_1$-distance (denoted as $\|\cdot\|_1$). Mathematically,
\begin{equation*}
\mathcal{L}_{recon} =
\mathbb{E}\left[ \left\| (\hat{y} - y) \right\|_1 \right].
\end{equation*}

\subsubsection{\textbf{Perceptual Loss:}}
To improve the visual fidelity in $\hat{y}$, a perceptual loss~\cite{johnson2016perceptual} is imposed. Assuming $\phi_{vgg}(\cdot) \in \mathbb{R}^{C_l \times H_l \times W_l}$ denote the spatial feature map extracted from the $l$-th layer of a pretrained VGG19~\cite{simonyan2014very} network, the perceptual loss is defined as
\begin{equation*}
\mathcal{L}_{perc} =
\sum_{l \in \mathbb{L}} 
\omega_l \, 
\left\|
\phi_{vgg}(\hat{y}) - \phi_{vgg}(y)
\right\|_1,
\end{equation*}
where $\mathbb{L} = \{3, 8, 17, 26\}$ denotes the selected layers of the VGG19 network, and $\omega_l=\{1.0, 0.75, 0.5, 0.25\}$ represents the corresponding weighing factors, to emphasize multi-level perceptual consistency with progressively reducing contributions from deeper layers.

\subsubsection{\textbf{Adversarial Loss:}}
To ensure realistic glyph generation, a hinge-based adversarial loss is used. Assuming $D_{adv}(\cdot)$ denotes the output from the discriminator, the adversarial loss for the discriminator network is defined as
\begin{equation*}
\begin{aligned}
\mathcal{L}_{adv}^{D} =
&\ \mathbb{E}_{y \sim p_{data}} 
\big[\max(0, 1 - D_{adv}(y))\big] \\
&+ \mathbb{E}_{\hat{y} \sim p_G} 
\big[\max(0, 1 + D_{adv}(\hat{y}))\big],
\end{aligned}
\end{equation*}

\noindent
and the adversarial loss for the generator network is defined as
\begin{equation*}
\mathcal{L}_{adv}^{G} =
- \mathbb{E}_{\hat{y} \sim p_G}
\left[ D_{adv}(\hat{y}) \right],
\end{equation*}

\noindent
where $p_{data}$ denotes the distribution of the real samples and $p_G$ denotes the distribution of the generated samples.

\subsubsection{\textbf{Auxillary Classification Loss:}}
Given auxiliary heads in $D$, $\theta_{style}(\cdot)$ and $\theta_{content}(\cdot)$, and ground-truth labels $y^{style}_{label}$ and $y^{content}_{label}$, the classification loss \cite{odena2017conditional} for the discriminator is defined as
\begin{equation*}
\mathcal{L}_{cls}^{D} =
\mathcal{L}_{CE}(\theta_{style}(y), y^{style}_{label})
+
\mathcal{L}_{CE}(\theta_{content}(y), y^{content}_{label}),
\end{equation*}

\noindent
and the classification loss \cite{odena2017conditional} for the generator is defined as
\begin{equation*}
\mathcal{L}_{cls}^{G} =
\mathcal{L}_{CE}(\theta_{style}(\hat{y}), y^{style}_{label})
+
\mathcal{L}_{CE}(\theta_{content}(\hat{y}), y^{content}_{label}),
\end{equation*}

\noindent
where $\mathcal{L}_{CE}$ denotes the cross-entropy loss.

\subsubsection{\textbf{Disentanglement Loss:}}
To ensure a well-structured embedding space, a contrastive loss is imposed to encourage compact clusters for positive pairs and large margins between negative pairs.

For the style features, given $e^{style}_{s}$ as the anchor embedding and the corresponding positive and negative style embeddings $e^{style}_{p}$ and $e^{style}_{n}$, the cosine similarity is computed for both positive $\{e^{style}_{s},e^{style}_{p}\}$ and negative $\{e^{style}_{s},e^{style}_{n}\}$ pairs, denoted as $s_{ps}$ and $s_{ns}$, respectively.

Circle Loss~\cite{sun2020circle} assigns adaptive weights $\delta_{ps}$ and $\delta_{ns}$ to positive and negative pairs, respectively, based on their relative optimization difficulty, with a margin hyperparameter $\eta$. During optimization, the adaptive weighing factors are treated as constants. The Circle Loss for a single triplet of style features is defined as
\begin{equation*}
\mathcal{L}_{circle}^{style} =
\log \left(
1 +
e^{
\gamma \left( 
\delta_{ns} (s_{ns} - \eta)
-
\delta_{ps} (s_{ps} - (1 - \eta))
\right)
}
\right),
\end{equation*}

\noindent
where $\gamma$ is a scaling factor controlling the strength of optimization. 

Similarly, for the content features, given positive $\{e^{content}_{s},e^{content}_{p}\}$ and negative $\{e^{content}_{s},e^{content}_{n}\}$ pairs, the cosine similarities $s_{pc}$ and $s_{nc}$, and the adaptive weighing factors $\delta_{pc}$ and $\delta_{nc}$, the Circle Loss for the content features is given by
\begin{equation*}
\mathcal{L}_{circle}^{content} =
\log \left(
1 +
e^{
\gamma \left( 
\delta_{nc} (s_{nc} - \eta)
-
\delta_{pc} (s_{pc} - (1 - \eta))
\right)
}
\right).
\end{equation*}

\noindent
Therefore, the cumulative disturbance loss can be defined as
\[
\mathcal{L}_{dist} = \mathcal{L}_{circle}^{style} + \mathcal{L}_{circle}^{content}.
\]

\subsubsection{\textbf{Latent Loss:}}
To further enhance structural and perceptual consistency, a latent reconstruction loss is imposed using a pretrained Stable Diffusion v2 VAE Encoder~\cite{rombach2022high}. Instead of constraining only image-level similarity, the generated image is aligned with the ground-truth image in the latent space.

Let $\phi_{SD}(\cdot)$ denote the encoder of the pretrained Stable Diffusion v2 VAE Encoder~\cite{rombach2022high}. Given the generated image $\hat{y}$ and the target image $y$, their corresponding latent representations $\phi_{SD}(\hat{y})$, and $\phi_{SD}(y)$; the latent loss is defined as
\begin{equation*}
\mathcal{L}_{latent} =
\left\|
\phi_{SD}(\hat{y}) - \phi_{SD}(y)
\right\|_1.
\end{equation*}

During training, the VAE encoder is kept frozen, and gradients are not propagated through $\phi_{SD}(\cdot)$. Minimizing $\mathcal{L}_{latent}$ encourages the generated glyph to match the target in the latent space, thereby enforcing intricate style and structural consistency beyond the generic pixel-level supervision.

\subsubsection{\textbf{Total Loss:}}
The overall objective follows a min–max optimization strategy between the generator $G$ and discriminator $D$, $\min_{G} \max_{D}\mathcal{L}(G, D)$, 
where the generator objective is defined as
\begin{align*}
\mathcal{L}_G =
&\;
\lambda_{recon}\mathcal{L}_{recon}
+ \lambda_{perc}\mathcal{L}_{perc}
+ \lambda_{dist}\mathcal{L}_{dist} \nonumber \\
&
+ \lambda_{latent}\mathcal{L}_{latent}
+ \lambda_{adv}\mathcal{L}_{adv}^G
+ \lambda_{cls}\mathcal{L}_{cls}^G,
\end{align*}

\noindent
and the discriminator objective is defined as
\begin{equation*}
\mathcal{L}_D =
\lambda_{adv}\mathcal{L}_{adv}^D
+ \lambda_{cls}\mathcal{L}_{cls}^D,
\end{equation*}

\noindent
where the $\lambda_\star$ values denote respective weighing factors for different loss components.

\section{Results}
\label{sec:results}

\begin{figure*}[t]
    \centering
    \includegraphics[width=\linewidth]{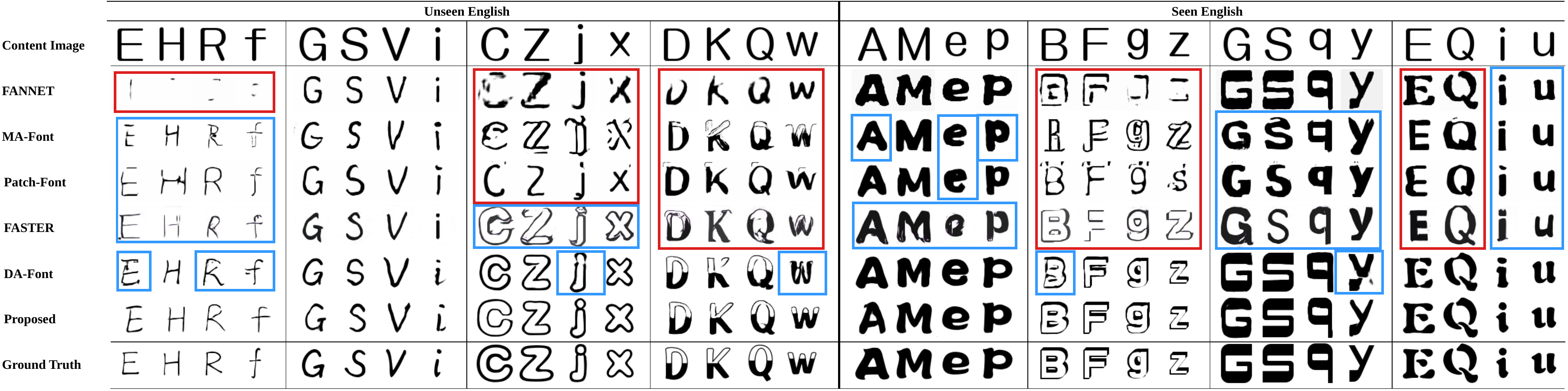}
    \caption{Qualitative comparison results on Unseen English and Seen English fonts. Boxes marked in \textcolor{red}{\emph{red}} highlight issues like failure of style adaptation, structural deformation, and missing strokes. Boxes marked in \textcolor{blue}{\emph{blue}} highlight issues like minor appearance of artifacts and minor structural inconsistencies.}
    \label{fig:english_results}
\end{figure*}

\begin{figure*}[t]
    \centering
    \includegraphics[width=\linewidth]{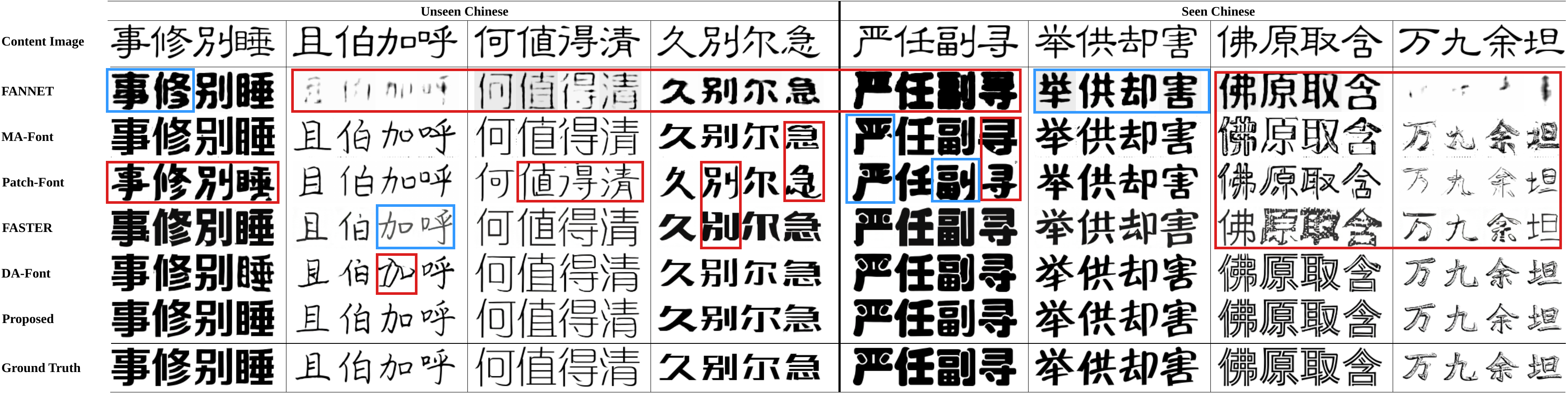}
    \caption{Qualitative comparison results on Unseen Chinese and Seen Chinese fonts. Boxes marked in \textcolor{red}{\emph{red}} highlight issues like failure of style adaptation, structural deformation, and missing strokes. Boxes marked in \textcolor{blue}{\emph{blue}} highlight issues like minor appearance of artifacts and minor structural inconsistencies.}
    \label{fig:chinese_results}
\end{figure*}

\begin{table*}[t]
\centering
\caption{Quantitative comparison of the proposed method with the existing state-of-the-art techniques on Unseen English and Seen English fonts. The best scores are shown in \textbf{bold} and the second best scores are \underline{underlined}.}
\label{tab:quantitative_compare_english}
\begin{tabular}{l cccc cccc c}

\toprule

& \multicolumn{4}{c}{Unseen English} & \multicolumn{4}{c}{Seen English} & \\
\cmidrule(lr){2-5}
\cmidrule(lr){6-9}

Method 
& L1 $\downarrow$ & RMSE $\downarrow$ & SSIM $\uparrow$ & LPIPS $\downarrow$
& L1 $\downarrow$ & RMSE $\downarrow$ & SSIM $\uparrow$ & LPIPS $\downarrow$
& User Study$\uparrow$ \\

\midrule

FANNET~\cite{roy2020stefann}
& 0.077 & \underline{0.239} & \underline{0.731} & 0.185
& 0.078 & 0.238 & \underline{0.779} & 0.116
& 10.789 \\

MA-Font~\cite{qiu2024ma}
& 0.089 & 0.272 & 0.694 & 0.143
& 0.087 & 0.269 & 0.712 & 0.138
& ~6.710 \\

PatchFont~\cite{memon2025patch}
& 0.098 & 0.266 & 0.675 & 0.143
& 0.100 & 0.281 & 0.680 & 0.143
& ~5.657 \\

FASTER~\cite{das2025faster} 
& 0.075 & 0.246 & 0.723 & 0.139
& 0.077 & 0.239 & 0.731 & 0.138
& ~7.894 \\

DA-Font~\cite{chen2025font} 
& \underline{0.074} & 0.243 & 0.713 & \underline{0.111}
& \underline{0.069} & \underline{0.223} & 0.775 & \underline{0.084}
& \underline{15.526} \\

\rowcolor[HTML]{CCFFCC}
Proposed 
& \textbf{0.072} & \textbf{0.237} & \textbf{0.739} & \textbf{0.108}
& \textbf{0.061} & \textbf{0.217} & \textbf{0.790} & \textbf{0.087}
& \textbf{53.421} \\

\bottomrule

\end{tabular}
\end{table*}

\begin{table*}[t]
\centering
\caption{Quantitative comparison of the proposed method with the existing state-of-the-art techniques on Unseen Chinese and Seen Chinese fonts. The best scores are shown in \textbf{bold} and the second best scores are \underline{underlined}.} 
\label{tab:quantitative_compare_chinese}
\begin{tabular}{l cccc cccc c}

\toprule

& \multicolumn{4}{c}{Unseen Chinese} & \multicolumn{4}{c}{Seen Chinese} & \\
\cmidrule(lr){2-5}
\cmidrule(lr){6-9}

Method 
& L1 $\downarrow$ & RMSE $\downarrow$ & SSIM $\uparrow$ & LPIPS $\downarrow$
& L1 $\downarrow$ & RMSE $\downarrow$ & SSIM $\uparrow$ & LPIPS $\downarrow$
& User Study$\uparrow$ \\

\midrule

FANNET~\cite{roy2020stefann} 
& \textbf{0.161} & \textbf{0.347} & 0.449 & 0.289
& 0.147 & 0.334 & 0.502 & 0.291
& ~2.105 \\

MA-Font~\cite{qiu2024ma} 
& 0.176 & 0.374 & 0.430 & 0.150
& 0.172 & 0.372 & 0.456 & 0.160
& ~9.605 \\

PatchFont~\cite{memon2025patch} 
& 0.251 & 0.442 & 0.288 & 0.224
& 0.225 & 0.420 & 0.353 & 0.204
& ~3.158 \\

FASTER~\cite{das2025faster} 
& 0.197 & 0.339 & 0.458 & 0.160
& \textbf{0.093} & \underline{0.297} & 0.484 & 0.166
& ~8.421 \\

DA-Font~\cite{chen2025font} 
& 0.166 & 0.357 & 0.469 & 0.143
& 0.125 & 0.303 & \underline{0.603} & \underline{0.107}
& \underline{21.053} \\

\rowcolor[HTML]{CCFFCC}
Proposed 
& \underline{0.162} & \underline{0.350} & \textbf{0.484} & \textbf{0.136}
& \underline{0.116} & \textbf{0.289} & \textbf{0.631} & \textbf{0.099}
& \textbf{55.658} \\

\bottomrule

\end{tabular}
\end{table*}

\subsection{Experimental Setup}

\subsubsection{\textbf{Dataset:}}
To evaluate the efficacy of the proposed pipeline, a multi-script glyph dataset~\cite{li2021few}, containing both Latin (for English) and Chinese characters, is used in the experimental studies. The English dataset consists of $811$ unique fonts, where $783$ fonts \textbf{(Seen English)} are used for training and the remaining $28$ fonts \textbf{(Unseen English)} for testing. Similarly, the Chinese dataset comprises $521$ fonts, of which $507$ \textbf{(Seen Chinese)} are used for training and the remaining $14$ \textbf{(Unseen Chinese)} for testing. The glyph samples include all $52$ English characters ($26$ uppercase + $26$ lowercase) and $993$ Chinese characters. Each glyph image in the dataset has a spatial resolution of $64 \times 64$.

\subsubsection{\textbf{Evaluation metrics:}}
The quantitative analysis uses four metrics to measure the quality of the generated samples. The pixel-level deviation between the generated samples and the ground truth is measured using \textbf{L1} distance and Root Mean Square Error \textbf{(RMSE)}. For perceptual comparison, Structural Similarity Index Measure \textbf{(SSIM)}~\cite{wang2004image} and Learned Perceptual Image Patch Similarity \textbf{(LPIPS)}~\cite{zhang2018unreasonable} are used. SSIM measures the similarity between the generated and real images by comparing luminance, contrast, and structure rather than only pixel-wise error. LPIPS evaluates perceptual similarity between two images by comparing the respective feature spaces using a pretrained deep neural network as the feature-extracting backbone.

As a quantifiable metric for visual quality is an open challenge in computer vision, the experimental analysis includes an opinion-based user study for a subjective visual quality assessment. The study uses a set of 20 randomly selected characters (10 English + 10 Chinese) with seven images for each character. Out of these seven images, one instance is the ground truth, and the remaining six images are generated using six different methods, including the proposed technique. During the study involving 76 individuals, the ground truth is shown to the user as a visual reference, and the user is tasked with selecting the visually closest image to the reference from the six possible options. The Mean Opinion Score \textbf{(MOS)} is evaluated as the average fraction of times one method is preferred over others.

\subsubsection{\textbf{Implementation details:}}
The proposed network is trained for $500$ epochs, with a batch size of $64$. The optimization uses the Adam~\cite{kingma2014adam} optimizer with a learning rate of $0.0002$. The embedding dimensions ($\in \mathbb{R}^{d}$) for both style and content representation are set to $768$, where each head ($\in \mathbb{R}^\frac{d}{3}$) has a dimension of 256. The weights of different loss terms $\lambda_{recon}$, $\lambda_{perc}$, $\lambda_{dist}$, $\lambda_{latent}$, $\lambda_{cls}$, and $\lambda_{adv}$ are set (using emperical study) to $5.0$, $1.0$, $0.2$, $0.15$, $1.0$, and $0.5$, respectively. For a font $f_a$, the cardinality of the set of available observations, $|\mathcal{O}_a|$ is set to $10$. The experiments are performed on a single Nvidia GeForce RTX $4080$ GPU with $16$GB VRAM.

\subsection{Results and Comparisons}
Multiple recent font generation approaches rely on predefined stroke representations. For a fair comparison, the proposed method is evaluated against existing strategies that do not rely on such decompositions. In this paper, the qualitative and quantitative studies compare the proposed method with existing SOTA techniques, including FANNET~\cite{roy2020stefann}, MA-Font~\cite{qiu2024ma}, PatchFont~\cite{memon2025patch}, FASTER~\cite{das2025faster}, and DA-Font~\cite{chen2025font}.

From Figure~\ref{fig:english_results} and Figure~\ref{fig:chinese_results}, it is evident that the proposed method achieves the best generation quality among the competing methods for both English and Chinese fonts. It is worth mentioning that the proposed method can capture complex font styles while maintaining structural consistency. Although FANNET can capture simple and thick fonts, it struggles with thin and complex patterns for both English and Chinese. MA-Font and PatchFont face challenges due to structural deformation, style inconsistency, and artifacts. Although FASTER yields comparatively better results than the last three, especially for Chinese, it still exhibits artifacts and structural deformities. DA-Font achieves higher visual quality, but structural deformations and artifacts persist for complex font styles.

Furthermore, Table~\ref{tab:quantitative_compare_english} and Table~\ref{tab:quantitative_compare_chinese} highlight the quantitative comparison with the SOTA methods, where the proposed method outperforms most of the metrics for both English and Chinese datasets. Notably, the proposed method achieves a significantly higher user score compared to other techniques, with users preferring the proposed method in 53.42\% and 55.66\% of cases for English and Chinese fonts, respectively. Overall, the users preferred the subjective generation quality of the proposed method in $54.54\%$ of cases, demonstrating superior generative capabilities over existing SOTA techniques.

\subsection{Ablation Studies}
To validate the efficacy of the network components and hyperparameters associated with the proposed pipeline, multiple ablation studies are conducted. All the ablation experiments are performed on the Unseen English fonts.

\subsubsection{\textbf{Impact of the RS Module:}}
\begin{figure}[h]
    \centering
    \includegraphics[width=\linewidth]{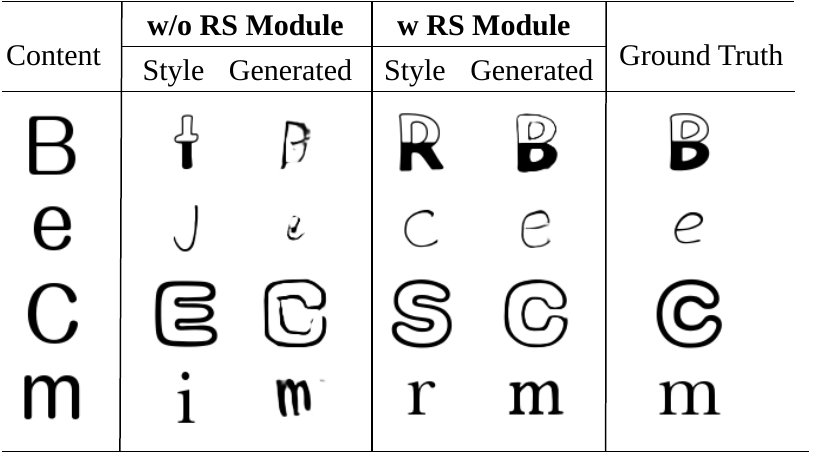}
    \caption{Qualitative comparison of the generation quality \emph{with} and \emph{without} (in both train and test phases) incorporating the RS Module into the proposed pipeline.}
    \label{fig:ablation_rsmodule}
    \vspace{-1.0em}
\end{figure}

To assess the contribution of the proposed RS Module, a study is conducted by training and evaluating the model with and without the RS Module under identical settings. As illustrated in Figure~\ref{fig:ablation_rsmodule}, incorporating the RS Module significantly improves generation quality. Specifically, the generated images with the RS Module exhibit sharper edges, clearer stroke boundaries, and improved structural alignment with the ground truth. In contrast, the model without the RS Module tends to produce comparatively blurred contours and structural inconsistencies, particularly in regions requiring precise style transfer. Furthermore, the RS Module enhances spatially localized style adaptation. By effectively leveraging the most relevant reference style features, it enables higher-quality feature aggregation at corresponding spatial regions. A quantitative analysis is reported in Table~\ref{tab:ablation_rsmodule}, showing that the inclusion of RS Module results in significant relative improvements of $26.58\%$, $17.29\%$, $9.31\%$, and $35.35\%$ in L1, RMSE, SSIM, and LPIPS metrics, respectively.

\begin{table}[h]
\centering
\caption{Quantitative analysis of the impact of the RS Module on the generation quality.}
\label{tab:ablation_rsmodule}
\setlength{\tabcolsep}{6pt}
\resizebox{\columnwidth}{!}{%
\begin{tabular}{cc cccc}
\toprule
\multicolumn{2}{c}{RS Module} & \multicolumn{4}{c}{Metrics} \\
\cmidrule(lr){1-2} \cmidrule(lr){3-6}
Training & Testing 
& L1 $\downarrow$ 
& RMSE $\downarrow$ 
& SSIM $\uparrow$ 
& LPIPS $\downarrow$ \\
\midrule
\xmark & \xmark & 0.098 & 0.287 & 0.676 & 0.167 \\
\xmark & \cmark & 0.085 & 0.266 & 0.699 & 0.141 \\
\cmark & \xmark & 0.081 & 0.256 & 0.713 & 0.129 \\
\rowcolor[HTML]{CCFFCC}
\cmark & \cmark & \textbf{0.072} & \textbf{0.237} & \textbf{0.739} & \textbf{0.108} \\
\bottomrule
\end{tabular}
}
\end{table}

\subsubsection{\textbf{Analysis of usage of various loss functions:}}
To improve generation quality, the proposed method is optimized using $\mathcal{L}_{cls}$ and $\mathcal{L}_{latent}$ along with $\mathcal{L}_{recon}$, $\mathcal{L}_{perc}$, $\mathcal{L}_{dist}$, and $\mathcal{L}_{adv}$. To justify their contribution, an ablation analysis is performed. As reported in Table~\ref{tab:ablation_loss}, jointly incorporating these losses results in relative improvements of $3.1\%$, $2.42\%$, $1.26\%$, and $4.17\%$ in L1, RMSE, SSIM, and LPIPS, respectively. Notably, the objective function of the proposed pipeline achieves the best performance across all metrics while adding $\mathcal{L}_{cls}$ and $\mathcal{L}_{latent}$, showing the efficacy of $\mathcal{L}_{cls}$ and $\mathcal{L}_{latent}$ in facilitating more effective parameter optimization and improved reconstruction fidelity.

\begin{table}[h]
\centering
\caption{Quantitative analysis of the effectiveness of $\mathcal{L}_{\text{cls}}$ and $\mathcal{L}_{\text{latent}}$ on the generation quality.}
\label{tab:ablation_loss}
\setlength{\tabcolsep}{6pt}
\resizebox{\columnwidth}{!}{%
\begin{tabular}{cc cccc}
\toprule
\multicolumn{2}{c}{Loss Function} & \multicolumn{4}{c}{Metrics} \\
\cmidrule(lr){1-2} \cmidrule(lr){3-6}
$\mathcal{L}_{\text{cls}}$ & $\mathcal{L}_{\text{latent}}$ 
& L1 $\downarrow$ 
& RMSE $\downarrow$ 
& SSIM $\uparrow$ 
& LPIPS $\downarrow$ \\
\midrule
\xmark & \xmark & 0.074 & 0.243 & 0.730 & 0.113 \\
\xmark & \cmark & 0.073 & 0.240 & 0.736 & 0.108 \\
\cmark & \xmark & 0.075 & 0.244 & 0.729 & 0.114 \\
\rowcolor[HTML]{CCFFCC}
\cmark & \cmark & \textbf{0.072} & \textbf{0.237} & \textbf{0.739} & \textbf{0.108} \\
\bottomrule
\end{tabular}
}
\end{table}

\subsubsection{\textbf{Analysis of Embedding Dimension:}}
To effectively bound the feature space for both font style and content, a study has been made by varying the embedding dimensionality of each head of the MSHB and the MCHB. Table~\ref{tab:ablation_dimension} shows that the model achieves the best performance when the embedding dimension of each head is set to $256$. Reducing the dimension to $128$ leads to performance degradation, due to insufficient representational capacity to capture discriminative style and structural features. Moreover, increasing the dimension to $512$ also degrades the performance, suggesting over-parameterization and potential overfitting.

\begin{table}[h]
\centering
\caption{Quantitative analysis of the effect of embedding dimension of each head ($\in \mathbb{R} ^\frac{d}{3}$) on the generation quality.}
\label{tab:ablation_dimension}
\resizebox{\columnwidth}{!}{%
\begin{tabular}{ccccc}
\toprule
Head Dimension & L1 $\downarrow$ & RMSE $\downarrow$ & SSIM $\uparrow$ & LPIPS $\downarrow$ \\
\midrule
128 & 0.074 & 0.241 & 0.734 & 0.108 \\
\rowcolor[HTML]{CCFFCC}
256 & \textbf{0.072} & \textbf{0.237} & \textbf{0.739} & \textbf{0.108} \\
512 & 0.074 & 0.241 & 0.734 & 0.111 \\
\bottomrule
\end{tabular}
}
\end{table}

\section{Conclusion}
\label{sec:conclusion}

In this paper, a novel font generation method, DRG-Font, is proposed, which dynamically selects the style reference by using a similarity measure to capture better local patterns. The proposed generative architecture adopts a contrasting learning strategy to learn the disentangled style and content latent spaces. The multiscale features captured through dedicated style and content heads of the encoder are used by the decoder for cross-feature aggregation, generating high-quality instances of the target glyph. The experimental results show the efficacy of the proposed method across English and Chinese glyphs, significantly outperforming the existing SOTA techniques.

{
    \small
    \bibliographystyle{ieeenat_fullname}
    \bibliography{main}
}


\end{document}